\documentclass{article}

\usepackage[main,preprint]{neurips_2026}

\usepackage[utf8]{inputenc}
\usepackage[T1]{fontenc}
\usepackage{microtype}
\usepackage{graphicx}
\usepackage{subfigure}
\usepackage{booktabs}
\usepackage{hyperref}
\usepackage{url}
\usepackage{amsmath}
\usepackage{amssymb}
\usepackage{mathtools}
\usepackage{amsthm}
\usepackage{xcolor}
\usepackage{algorithm}
\usepackage{algorithmic}
\usepackage[many]{tcolorbox}
\usepackage{capt-of}
\tcbuselibrary{listings}

\newtcblisting{mylisting}{
  listing only,
  hbox,
  listing options={
    basicstyle=\small\ttfamily
  },
}

\DeclareMathOperator*{\argmax}{arg\,max}
\definecolor{mybgcolor}{HTML}{F0F0F0}

\usepackage[capitalize,noabbrev]{cleveref}

\theoremstyle{plain}
\newtheorem{theorem}{Theorem}[section]

\theoremstyle{definition}
\newtheorem{definition}[theorem]{Definition}

\theoremstyle{remark}

\usepackage[disable,textsize=tiny]{todonotes}

\title{SwiftMem: Fast Agentic Memory via Query-aware Indexing}

\author{%
  \begin{tabular}{c}
    \textbf{Anxin Tian} \quad \textbf{Yiming Li} \quad \textbf{Xing Li} \quad \textbf{Hui-Ling Zhen} \quad \textbf{Lei Chen} \\
    \textbf{Xianzhi Yu} \quad \textbf{Zhenhua Dong} \quad \textbf{Mingxuan Yuan}\thanks{yuan.mingxuan@huawei.com}
  \end{tabular} \\
  Huawei, Hong Kong \\
}

\begin{document}

\maketitle

\begin{abstract}
Agentic memory systems have become critical for enabling LLM agents to maintain long-term context and retrieve relevant information efficiently.
However, existing memory frameworks often perform query-agnostic retrieval over the full memory embedding space even when their storage layer is backed by efficient vector indexes such as HNSW.
This full-scope retrieval path creates latency bottlenecks as memory grows, hindering real-time agent interactions.
We propose SwiftMem, a query-aware agentic memory system that narrows retrieval to query-relevant memory subsets through specialized indexing over temporal and semantic dimensions.
Our temporal index enables logarithmic-time range queries for time-sensitive retrieval, while the semantic DAG-Tag index maps queries to relevant topics through hierarchical tag structures.
To address memory fragmentation during growth, we introduce an embedding-tag co-consolidation mechanism that reorganizes storage based on semantic clusters to improve locality.
Across LoCoMo and LongMemEval$_S$, SwiftMem reaches 10.8/12.7 ms search latency while maintaining competitive LLM-judge accuracy against strong HNSW-backed memory systems.
On the calibrated benchmark, LoCoMo Refined, SwiftMem remains close to the top LLM-judge score while preserving an order-of-magnitude latency advantage.
The code is available at \url{https://github.com/EdwardTex/SwiftMem}.
\end{abstract}

\section{Introduction}
\label{sec:intro}

The emergence of agentic AI systems has transformed how large language models (LLMs) interact with users, evolving from short question-answering sessions to long-running task execution and tool use~\cite{su2025scalingagentscontinualpretraining,zhai2025agentevolverefficientselfevolvingagent,lumer2025memtooloptimizingshorttermmemory,li2025loopserveadaptivedualphasellm}.
Long-term memory is central to this evolution: agents must store, organize, and retrieve relevant information from extended conversational histories~\cite{chhikara2025mem0,qian2025memoragboostinglongcontext,fang2025lightmemlightweightefficientmemoryaugmented}.
As interactions span thousands of turns, retrieval latency becomes a user-facing bottleneck.

Current memory frameworks~\cite{xu2025amemagenticmemoryllm,zhang2025gmemorytracinghierarchicalmemory,nan2025nemori} typically sit above storage systems such as vector databases, relational stores, and knowledge graphs.
These systems can already use strong approximate nearest-neighbor (ANN) indexes, such as HNSW, IVF-style indexes, or DiskANN-like graph indexes~\cite{malkov2020hnsw,subramanya2019diskann,douze2024faiss,2021milvus}.
Our critique is therefore not that practical baselines perform an unoptimized linear scan.
Rather, their retrieval path is usually \emph{query-agnostic}: each query is routed to a broad search over the full indexed memory embedding space, even when the query itself specifies a narrow temporal interval or semantic topic.
ANN indexes reduce the cost of searching a given candidate universe, but they do not decide which memory region should be searched.

\begin{figure*}[t]
    \centering
    \includegraphics[width=0.7\linewidth]{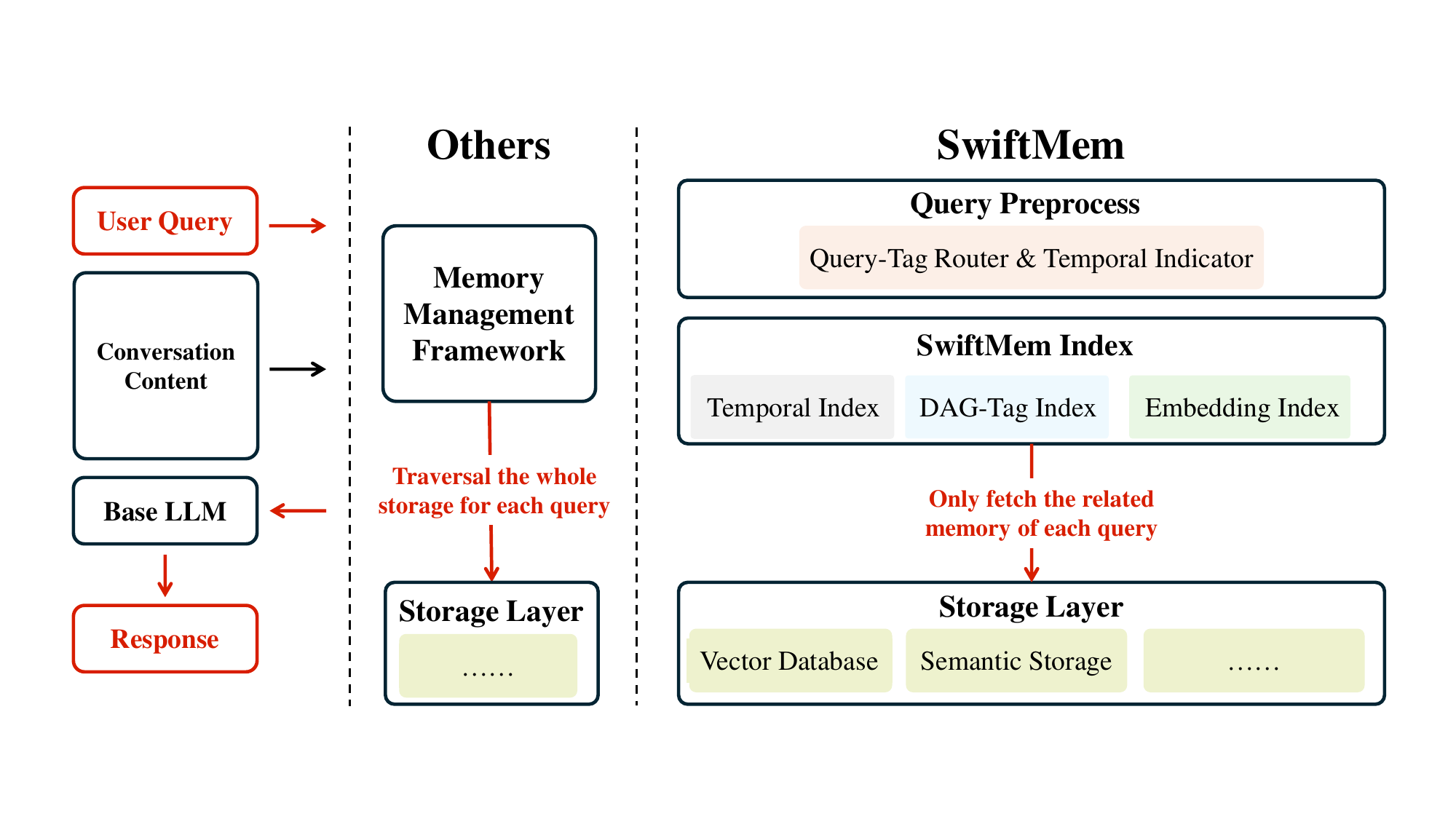}
    % \vspace{-20pt}
    \caption{Architectural comparison between query-agnostic memory retrieval and SwiftMem. Existing systems can use ANN-backed vector search, but still query the broad memory space. SwiftMem first routes the query through temporal and semantic indexes to select a smaller relevant subset.}
    \label{fig:comparison}
\end{figure*}

\begin{figure}[t]
    \centering
    \includegraphics[width=0.85\linewidth]{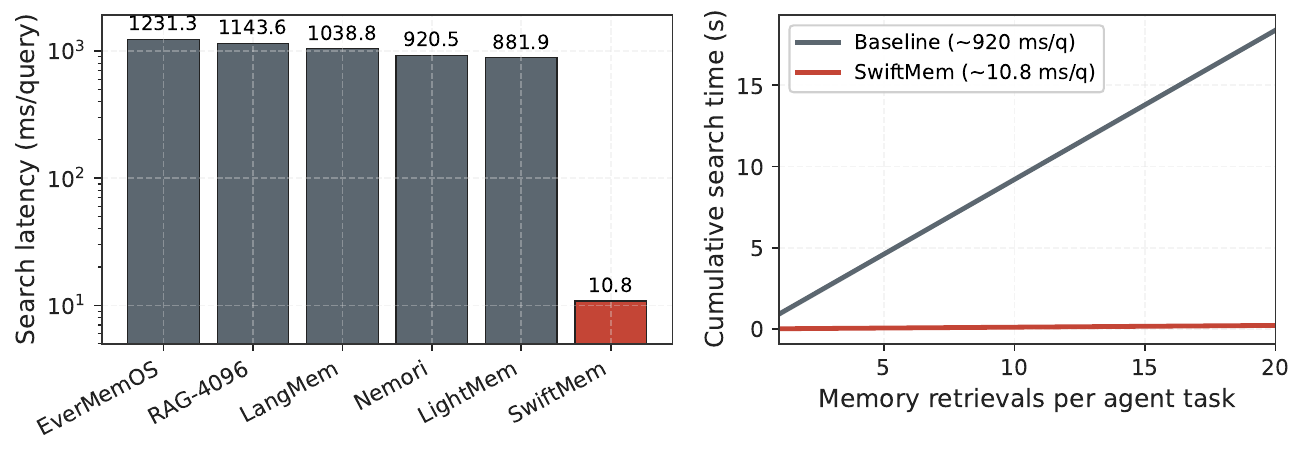}
    % \vspace{-10pt}
    \caption{Empirical per-query search latency on LoCoMo for existing SOTA memory systems (left, log scale), and an illustrative cumulative search cost when an agent performs multiple memory retrievals within one task (right; the baseline curve uses $\sim$920\,ms/query as a representative strong memory-system order-of-magnitude).}
    \label{fig:motivation_latency}
\end{figure}

This distinction matters for real-time agents as illustrated in Figure~\ref{fig:motivation_latency}.
Under a unified rerun protocol on LoCoMo, strong memory baselines backed by vector retrieval still spend hundreds of milliseconds to more than one second per query, while LongMemEval increases the latency further as the memory context grows.
The inefficiency stems from ignoring two forms of locality that are common in conversational memory queries.
\textbf{Temporal locality} appears in explicit or implicit references such as ``What did we discuss last week?'' or ``When did I mention camping?''.
\textbf{Semantic locality} appears when the answer is tied to a small set of topics, entities, or preferences rather than the entire memory base.

We propose \textbf{SwiftMem}, a query-aware agentic memory system that narrows retrieval to query-relevant memory subsets before invoking the storage layer.
SwiftMem follows a simple principle: analyze the query to identify what should be searched, then search only where necessary.
This principle is realized through three components.
\textbf{(1) Temporal Index.} SwiftMem builds user-specific sorted timelines and global episode lookup, enabling logarithmic-time range queries for time-sensitive retrieval.
\textbf{(2) Semantic DAG-Tag Index.} SwiftMem organizes LLM-generated tags in a directed acyclic graph and routes each query to a bounded set of relevant tags in $O(k(\log |V| + D_{max}))$ time, where $k$ is the number of seed tags and $D_{max}$ is the bounded expansion depth.
\textbf{(3) Embedding Index with Co-consolidation.} SwiftMem keeps an embedding index for similarity retrieval and periodically reorganizes embeddings by semantic tag clusters to improve locality when vector search is activated.

\begin{figure*}[t]
    \centering
    \includegraphics[width=0.7\linewidth]{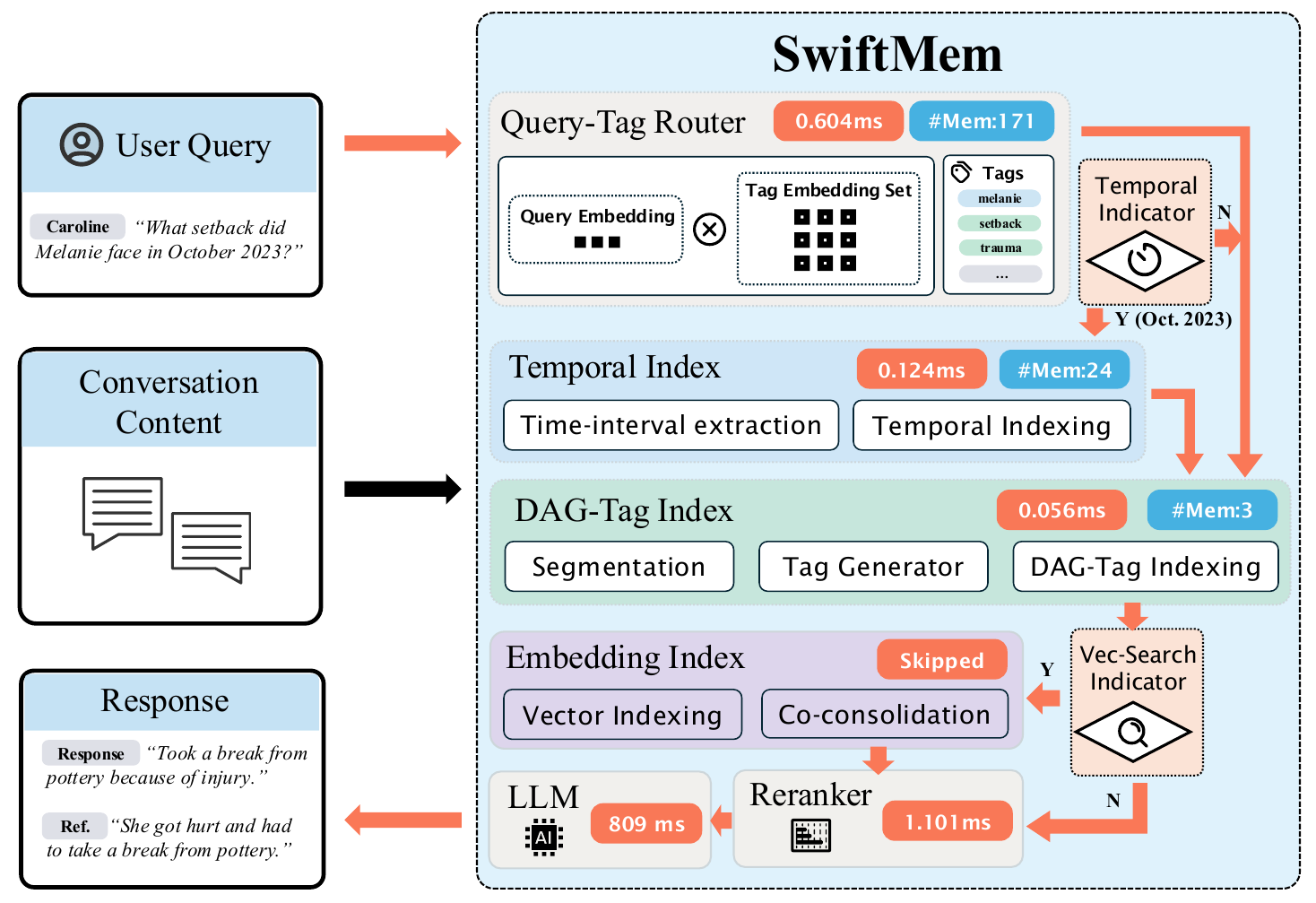}
    \caption{SwiftMem's end-to-end query workflow illustrated with a LoCoMo example~\cite{maharana2024evaluating}. The DAG-Tag index is the default retrieval backbone; temporal and embedding indexes are activated when the query requires them.}
    \label{fig:workflow}
\end{figure*}

We evaluate SwiftMem on LoCoMo~\cite{maharana2024evaluating}, LongMemEval~\cite{wulongmemeval} and LoCoMo Refined~\cite{locomorefined2026} under a unified harness.
The results show that SwiftMem achieves substantially lower search latency than ANN-backed memory baselines while preserving competitive LLM-judge accuracy and strong overlap scores (F1/BLEU-1).

\section{Related Work}
\label{sec:rw}

\subsection{Content-as-Memory}
\label{sec:cam}

Content-as-Memory (CaM) systems extract and store factual knowledge from conversational interactions, enabling agents to recall user preferences and contextual information across sessions.
Early work such as MemoryBank~\cite{zhong2023memorybankenhancinglargelanguage} introduced selective memory preservation inspired by cognitive science.
Recent systems explore diverse memory organizations: Zep~\cite{rasmussen2025zep} employs temporal knowledge graphs; Mem0~\cite{chhikara2025mem0} uses graph-based personalized memory extraction; MIRIX~\cite{wang2025mirixmultiagentmemoryllmbased} introduces multi-type memory architectures; Nemori~\cite{nan2025nemori} aligns episodic and semantic memory; LangMem~\cite{langchain2024} provides tooling with LangGraph integration; MemOS~\cite{li2025memosoperatingmemoryaugmentedgeneration} treats memory as an operational resource; MemoRAG~\cite{qian2025memoragboostinglongcontext} uses dual-system global retrieval; and LightMem~\cite{fang2025lightmemlightweightefficientmemoryaugmented} organizes memory into sensory, short-term, and long-term stages.

Recent works have begun addressing the efficiency challenge through structured retrieval.
SimpleMem~\cite{liu2026simplememefficientlifelongmemory} introduces semantic compression with query-aware retrieval but lacks hierarchical indexing structures.
Membox~\cite{tao2026memboxweavingtopiccontinuity} groups conversational turns by topic for continuity but uses a flat clustering structure.
LightMem is especially relevant because it also targets efficient long-term agent memory; however, it primarily reduces memory construction cost through pre-compression, buffering, and offline updates.
SwiftMem instead focuses on retrieval-time indexing: temporal routing, DAG-Tag routing, and co-consolidated embedding locality narrow the search space at query time.

\subsection{ANN Indexes for Vector Retrieval}
\label{sec:ann_related}

Approximate nearest-neighbor search is a mature area with strong indexes such as HNSW~\cite{malkov2020hnsw}, IVF/PQ as implemented in FAISS~\cite{douze2024faiss}, and DiskANN~\cite{subramanya2019diskann}.
Vector databases such as Milvus~\cite{2021milvus} expose these indexing families in practical systems.
ANN-Benchmarks~\cite{annbenchmarks} further shows that no single index dominates all datasets and metrics; HNSW is a strong and widely used option, while IVF and DiskANN offer different recall, throughput, memory, and build-time trade-offs.
SwiftMem is complementary to these indexes.
It does not claim that baselines use linear scan; instead, it reduces the candidate universe that an ANN-backed memory retriever must search by using temporal and semantic query structure.

\subsection{Trajectory-as-Memory}
\label{sec:tam}

Trajectory-as-Memory (TaM) systems focus on procedural learning.
Synapse~\cite{zheng2024synapsetrajectoryasexemplarpromptingmemory} pioneered trajectory-as-exemplar prompting; ReasoningBank~\cite{ouyang2025reasoningbankscalingagentselfevolving} distills reasoning strategies from self-judged experiences; AgentFold~\cite{ye2025agentfoldlonghorizonwebagents} introduces proactive context consolidation; MemGen~\cite{zhang2025memgenweavinggenerativelatent}, AgentEvolver~\cite{zhai2025agentevolverefficientselfevolvingagent}, Explore-to-Evolve~\cite{wang2025exploreevolvescalingevolved}, Scaling-Agents-via-CPT~\cite{su2025scalingagentscontinualpretraining}, and ATLAS~\cite{chen2025atlasagenttuninglearning} explore experience-driven evolution and trajectory-based learning.
While TaM works excel at procedural learning, they focus on trajectory utilization rather than query-time retrieval efficiency over long factual memory.

\section{Methodology}
\label{sec:method}

In this section, we first introduce the preliminaries for agentic memory systems. Then we introduce our index design based on temporal and semantic tags, and demonstrate how these enable rapid retrieval. 
Finally, we present the embedding-tag co-consolidation mechanism introduced to handle memory growth.

\subsection{Preliminaries}
\label{sec:prel}

We formalize the memory and retrieval setting that motivates SwiftMem.

\textbf{Memory Structures.}
Following cognitive psychology principles and established agentic memory frameworks~\cite{wang2025mirixmultiagentmemoryllmbased,nan2025nemori}, we adopt a multi-tier memory hierarchy including episodic memory and semantic memory; details appear in Appendix~\ref{app:mem_st}.

\textbf{Storage and Retrieval.}
Modern agentic memory systems often combine vector databases for episode embeddings $\{\mathbf{x}_1, \ldots, \mathbf{x}_{N_{mem}}\}$, relational databases for metadata, and graph stores for semantic relationships; see Appendix~\ref{app:sto_retri}.
The dense retrieval component can be backed by ANN indexes such as HNSW, IVF, or DiskANN~\cite{malkov2020hnsw,subramanya2019diskann,douze2024faiss}.
The bottleneck studied here is therefore not the absence of vector indexing, but the scope of retrieval: query-agnostic systems typically search over the full indexed memory embedding space, while many queries only require a temporally or semantically localized subset.
SwiftMem targets this missing routing layer by selecting the relevant memory region before ANN-backed embedding retrieval or reranking.

\subsection{Semantic DAG-Tag Index}
\label{sec:dag_tag}

SwiftMem addresses the retrieval latency issue through \textit{query-aware indexing} that analyzes query characteristics to identify relevant memory subsets \textit{before} similarity computation.
We propose a novel hierarchical tag indexing and routing mechanism that efficiently maps queries semantically to relevant tags through a Directed Acyclic Graph (DAG) structure. 
Our theoretical analysis demonstrates superior query efficiency compared to existing SOTA works.

\subsubsection{LLM-based Tag Generation}

The first stage leverages LLMs to analyze episode content and generate meaningful tags through carefully crafted prompts. For each episode, the LLM extracts 3-8 standardized tags (lowercase with underscores for multi-word concepts, e.g., `machine\_learning') that capture the main topics, themes, and contexts of the conversation. 
% Critically, SwiftMem instructs the LLM to identify hierarchical relationships between generated tags—for instance, recognizing `programming' as a broader concept encompassing `python\_coding'—which enables building a semantically meaningful tag structure rather than flat keyword lists.

The prompt engineering (see Appendix~\ref{sec:append}) focuses on four semantic dimensions to capture conversational essence: (1) topics and activities (e.g., `travel', `programming') representing primary themes; (2) locations and entities (e.g., `paris', `university') grounding conversations in concrete contexts; (3) emotions and intents (e.g., `career\_planning', `self\_acceptance') reflecting underlying motivations; and (4) specific over generic concepts (e.g., `italian\_cuisine' rather than `food') ensuring fine-grained semantic representation.
The system enforces strict formatting rules to maintain consistency—all tags lowercase, multi-word tags connected with underscores, and overly broad terms like `conversation' excluded.
These dimensions cover common semantic aspects in conversational contexts, though they may not be exhaustive for all dialogue scenarios.
% To ensure robustness, SwiftMem implements a fallback mechanism using embedding-based similarity when LLM generation fails.

\subsubsection{DAG-based Tag Index}
\label{sec:tag}

The core of SwiftMem is a hierarchical tag organization system that maintains semantic relationships while ensuring efficient retrieval. 
We formalize our approach as follows:

\begin{definition}[DAG-Tag Node Structure]
Let $\mathcal{G} = (V, E)$ be a directed acyclic graph where $V$ represents the set of tag nodes and $E$ represents the semantic relationships. Each node $v \in V$ is defined as:
\begin{equation}
v = (t, \mathcal{E}, \mathcal{P}, \mathcal{C}, \mathbf{e})
\end{equation}
where $t$ is the tag identifier, $\mathcal{E}$ is the episode set, $\mathcal{P}$ and $\mathcal{C}$ are parent and child tag sets respectively, and $\mathbf{e} \in \mathbb{R}^d$ is the tag's semantic embedding.
\end{definition}

\begin{theorem}[Semantic Specificity Hierarchy]
For any path $p = (v_1, ..., v_k)$ in $\mathcal{G}$, let $\mathcal{S}(v_i)$ denote the semantic specificity of node $v_i$. The specificity increases monotonically along the path:
\begin{equation}
\forall i < j: \mathcal{S}(v_i) < \mathcal{S}(v_j)
\end{equation}
\label{th:2}
\end{theorem}

This theorem formalizes the fundamental property of SwiftMem's hierarchical tag organization: semantic specificity strictly increases from parent to child nodes along any directed path in the DAG, where the $S(v)$ is operated via LLM-based pairwise scoring, prompting the model to rate how much more specific tag $v$ is compared to its parent (See Appendix~\ref{app:sem_spec}).
When integrating episode-level tags into the global DAG, we ensure the acyclic property through reachability checking: before adding edge $(u \to v)$, we verify that $u$ is not already an ancestor of $v$ in the existing DAG. 
SwiftMem implements an intelligent tag expansion mechanism that enriches search queries through breadth-first traversal of the tag DAG hierarchy. 
For each input tag, the system progressively explores more specific concepts level by level up to a specified depth parameter, enabling flexible search scope control. 
The example is shown in Appendix~\ref{sec:tag_ex}.

\subsubsection{Query-Tag Router}

We introduce a query-tag routing mechanism that efficiently maps natural language queries to relevant tags through semantic embedding alignment. The approach consists of three components:

\textbf{Embedding-based Alignment.} For a query $q$ with embedding $\mathbf{e}_q \in \mathbb{R}^{d}$ and a tag $t$ with embedding $\mathbf{x}_t \in \mathbb{R}^{d}$, we compute semantic similarity using cosine distance:
\begin{equation}
s(q, t) = \frac{\mathbf{e}_q \cdot \mathbf{x}_t}{||\mathbf{e}_q|| \cdot ||\mathbf{x}_t||}
\end{equation}
This embedding-based similarity enables the router to identify relevant tags even when queries use semantically related but lexically different terms.

\textbf{Tag Selection.} Given the tag set $V$ in the DAG structure, we select the top-$k$ most relevant tags to form the initial query tag set:
\begin{equation}
T_q = \argmax_{T' \subseteq V, |T'|=k} \sum_{t \in T'} s(q, t)
\end{equation}
where $|T_q| = k$ represents the number of selected tags that maximize cumulative semantic similarity with the query.

\textbf{Tag Expansion.} The selected tags $T_q$ are then expanded through the DAG hierarchy up to depth $D_{max}$, incorporating semantically related parent and child tags to enrich retrieval coverage while maintaining semantic coherence.
Expansion is not an unconstrained full-width traversal: SwiftMem maintains a score-keyed frontier heap and fixed frontier/candidate budgets, so the number of visited nodes per seed and per depth level is bounded in the query path.

% \begin{theorem}[DAG-Tag Query Complexity]
% Given a DAG-Tag index $\mathcal{G} = (V, E)$ and a query $q$ with initial tag set $T_q$ where $|T_q| = k$, let $|V|$ denote the total number of tags and $D_{max}$ be the maximum expansion depth. The time and space complexity are bounded by:
% \begin{equation}
% \begin{split}
% T_{query} & = O(k \cdot (\log |V| + D_{max})) \\
% S_{index} & = O(|V| \cdot (d + |\mathcal{P}_{avg}| + |\mathcal{C}_{avg}|))
% \end{split}
% \end{equation}
% where $d$ is the embedding dimension, and $|\mathcal{P}_{avg}|$, $|\mathcal{C}_{avg}|$ represent average parent and child set sizes.
% \end{theorem}

% The theorem establishes SwiftMem's efficiency advantages over traditional memory frameworks requiring $O(N_{mem})$ exhaustive search across all memory entries. SwiftMem achieves $O(k \cdot (\log |V| + D_{max}))$ query time through two optimizations: tag similarity computation in $O(k \cdot \log |V|)$ via efficient indexing, and hierarchical expansion in $O(k \cdot D_{max})$ by selective traversal. Since $k \ll |V| \ll N_{mem}$ and $D_{max} \ll |V|$ in practice, this represents orders of magnitude speedup. The space overhead $O(|V| \cdot (d + |\mathcal{P}_{avg}| + |\mathcal{C}_{avg}|))$ for maintaining tag embeddings and hierarchical relationships is modest compared to the query acceleration gained, enabling real-time semantic retrieval in large-scale conversational systems.

\begin{theorem}[DAG-Tag Query Complexity]
Given a DAG-Tag index $\mathcal{G} = (V, E)$ and a query $q$ with initial tag set $T_q$ where $|T_q| = k$, let $|V|$ denote the total number of tags and $D_{max}$ be the maximum expansion depth. The time and space complexity are bounded by:
\begin{equation}
\begin{split}
T_{query} & = O(k \cdot (\log |V| + D_{max})) \\
S_{index} & = O(|V| \cdot (d + |\mathcal{P}_{avg}| + |\mathcal{C}_{avg}|))
\end{split}
\end{equation}
where $d$ is the embedding dimension, and $|\mathcal{P}_{avg}|$, $|\mathcal{C}_{avg}|$ represent average parent and child set sizes.
\label{th:1}
\end{theorem}

The theorem is stated for this bounded-frontier operation rather than arbitrary DAG expansion.
The $O(k\log |V|)$ term comes from seed-tag lookup in the tag-embedding index, and the $O(kD_{max})$ term comes from bounded expansion over adjacency lists.
The full proof sketch and data-structure mapping are shown in Appendix~\ref{app:query}.

% \textbf{Analysis.} The theorem establishes SwiftMem's efficiency advantages over traditional memory frameworks requiring $O(N_{mem})$ exhaustive search across all memory entries. SwiftMem achieves $O(k \cdot (\log |V| + D_{max}))$ query time through two mechanisms: (1) tag similarity computation in $O(k \cdot \log |V|)$ via efficient nearest-neighbor search in the tag embedding space, and (2) hierarchical expansion in $O(k \cdot D_{max})$ by traversing from matched tags to their ancestors and descendants. Since $k \ll |V| \ll N_{mem}$ and $D_{max} \ll |V|$ in practice (e.g., $k \approx 3$-$10$, $D_{max} \approx 1$-$3$ for typical conversational taxonomies), this represents sub-linear scaling relative to memory size. The space overhead $O(|V| \cdot (d + |\mathcal{P}_{avg}| + |\mathcal{C}_{avg}|))$ for maintaining tag embeddings and DAG adjacency lists grows with semantic vocabulary rather than conversation length, remaining modest even for large-scale deployments. 
% The worst case occurs when queries lack semantic specificity, causing $T_q$ to match all tags in $V$ and triggering full DAG traversal. In this scenario, $T_{worst} = O(|V| \cdot D_{max})$, which degenerates toward baseline performance but does not exceed it. Typically real-world datasets show that most of queries exhibit semantic focus.

\subsection{Temporal Index}
\label{sec:temp}

During LLM Agent interactions, episodic memories are naturally generated and stored in chronological order. 
% The human brain similarly uses temporal order as a major index for memory retrieval, often recalling information through relative time references and maintaining strong temporal impressions of significant events. 
Many user queries for memory retrieval are inherently temporal, a fact that has been overlooked by many existing agentic memory works. 
While latest temporal indexing techniques~\cite{hou2024aeong,tian2024efficient} for specific domains, they are not suitable for the conversational memory retrieval in agentic scenarios. 

In our approach, we activate only the relevant memory subsets for temporal queries rather than performing query-agnostic retrieval across the entire memory store.

% \subsubsection{Query with Explicit Time Intervals}
% \label{sec:temp_ex}

For queries containing explicit time intervals, the optimal retrieval method is to directly extract timestamp information and locate the target episodes. 
% For instance, considering the query in Figure~\ref{fig:eg_extemp} containing a precise time interval (March 16, 2022), existing works perform crude temporal filtering using \texttt{{SELECT-WHERE}} clauses in SQL backends. 
Considering an explicit temporal query from LoCoMo~\cite{maharana2024evaluating}: ``Which recreational activity was James pursuing on March 16, 2022? (answer: bowling)'', it contains a precise time interval (March 16, 2022), existing works perform crude temporal filtering using \texttt{{SELECT-WHERE}} clauses in SQL backends.
In contrast, our approach constructs a temporal index following the natural order of memory growth, supporting queries across \textit{multiple time intervals}.

% \begin{figure}
%     \centering
%     \begin{minted}[
%     bgcolor=mybgcolor, % 背景色
%     frame=single,  % 边框
%     framesep=2mm,  % 边框与代码的距离
%     breaklines=true,
%     fontsize=\tiny % 字体大小
%     ]{json}
% "question": "Which recreational activity was James pursuing on March 16, 2022?",
% "answer": "bowling"
%     \end{minted}
%     \vspace{-10pt}
%     \caption{An explicit temporal query from LoCoMo~\cite{maharana2024evaluating}}
%     \label{fig:eg_extemp}
% \end{figure}

\textbf{Temporal Indexing.}
Unlike traditional temporal filtering in existing works, we build a temporal index that combines user-specific sorted timelines with global episodic lookup:

\begin{definition}[Temporal Index]
The temporal index $\mathcal{T}$ consists of: (1) User-specific sorted timelines: $\mathcal{L}_u = \{(t_i, e_i)\}$ where $t_i$ is timestamp and $e_i$ is episode ID. (2) Global episode lookup: $\mathcal{M}: e \rightarrow (u, t_i)$ mapping episodes to (user, timestamp) pairs.
\end{definition}

This index achieves $O(\log N_{mem})$ time complexity for temporal range queries through binary search on sorted timelines, while maintaining $O(1)$ episode metadata access through direct mapping $\mathcal{M}$. 
The structure supports efficient parallel processing across different user contexts and enables fast temporal operations including recent episode retrieval and temporal joins—operations that require complex and inefficient sequential scans in existing frameworks. 
Binary search-based insertion maintains strict temporal ordering within each user's timeline with $O(\log N_{mem})$ complexity for both insertions and queries.

\textbf{Temporal Indicator. }
Many queries involve temporal reasoning but differ from the explicit `time-to-event' mapping discussed above. 
Instead, they require `event-to-time' mapping, these queries cannot directly yield time intervals. 
The temporal indicator identifies whether a query contains explicit temporal intervals to route it to temporal indexing; otherwise, queries are routed directly to DAG-tag retrieval.
For such cases, see Appendix~\ref{app:temp_im} for details.

\subsection{Embedding Index}
\label{sec:coco}

\textbf{Vector Search Indicator.}
To avoid unnecessary embedding computations, we employ a selective vector search mechanism that activates only when DAG-Tag retrieval returns excessive candidates, indicating that tag-based filtering alone is insufficient for precise ranking.

\textbf{Co-consolidation Mechanism.}
To optimize vector search efficiency when activated, we propose a co-consolidation strategy that reorganizes embedding storage based on semantic tag clusters derived from the DAG-Tag structure.
The key insight is that tags sharing hierarchical relationships or frequently co-occurring in episodes tend to be queried together, making their embeddings good candidates for spatial locality optimization.

We define a tag cluster $C = (I_c, V_c, t_c, s_c)$ where $I_c$ is the cluster identifier, $V_c$ is the member tag set, $t_c$ is the centroid tag, and $s_c \in [0,1]$ is the cohesion score measuring intra-cluster connectivity (detailed in Appendix~\ref{app:clustering}).
Based on these clusters, we reorganize vector blocks so embeddings belonging to the same cluster are stored explicitly together, improving cache locality during retrieval.
We maintain a layout map tracking each tag's storage location; the consolidation process runs periodically with linear complexity, balancing cluster cohesion, size, and memory fragmentation through a scoring function (detailed in Appendix~\ref{app:consolidation}).

\section{Experiments}
\label{sec:exp}

We evaluate whether SwiftMem's query-aware routing, which narrows the candidate subset before invoking embedding retrieval or reranking, can break this latency barrier without compromising answer quality. 
Our experiments conduct a unified rerun against strong ANN-backed memory baselines and address three research questions: \textbf{RQ1}: Does SwiftMem preserve answer quality under consistent evaluation protocols? \textbf{RQ2}: What magnitude of query-time latency reduction does it achieve relative to comparable vector-indexed baselines? \textbf{RQ3}: How robust are these findings across open-weight models, ablations, and diagnostic stress tests?

\subsection{Experimental Setup}
\label{sec:exp_sp}

\textbf{Benchmarks.} We focus on three conversational-memory benchmarks: LoCoMo~\cite{maharana2024evaluating}, LongMemEval$_S$\cite{wulongmemeval}, and LoCoMo refined\cite{locomorefined2026}. LoCoMo comprises 10 multi-session dialogues ($\sim$24K tokens each) with 1,540 evaluation queries spanning multi-hop, temporal, open-domain, and single-hop reasoning. LongMemEval$_S$ contains 500 longer dialogues ($\sim$105K tokens each). LoCoMo refined recalibrates the original with stricter LLM judging and cleaned data for stronger reliability. 
% Full LongMemEval$_S$ results are in Appendix~\ref{app:longmem_results}.
Detailed workload statistics are in Appendix~\ref{app:workload}.

\textbf{Metrics.}
Following prior work~\cite{nan2025nemori,chhikara2025mem0}, we report LLM-as-Judge (LJ), F1, and BLEU-1 (B1) for answer quality.
We also report search latency per query in milliseconds and add-stage time in seconds when available. Best per column is bold and second best underlined.

\textbf{Baselines and ANN setting.}
We compare with FullContext, RAG-4096, LangMem~\cite{langchain2024}, Nemori~\cite{nan2025nemori}, LightMem~\cite{fang2025lightmemlightweightefficientmemoryaugmented}, and EverMemOS~\cite{hu2026evermemos}.
For methods with dense retrieval, the vector backend uses a vector index rather than an unoptimized linear scan; in our implementation this is HNSW-backed vector retrieval, a strong and widely used ANN choice~\cite{malkov2020hnsw,annbenchmarks}.
% Thus the comparison is not ANN versus linear scan.
% It compares \emph{query-agnostic full-memory retrieval} against SwiftMem's query-aware routing, which narrows the candidate memory subset before ANN-backed embedding retrieval or reranking.
Note that ANN-Benchmarks~\cite{annbenchmarks} further shows that HNSW, IVF, DiskANN, and other indexes trade off recall, throughput, memory, and build time across datasets, with no single golden-best index.

\subsection{Overall Quality--Latency Tradeoff}
\label{sec:exp_overall}

\begin{figure*}[t]
    \centering
    \includegraphics[width=\textwidth]{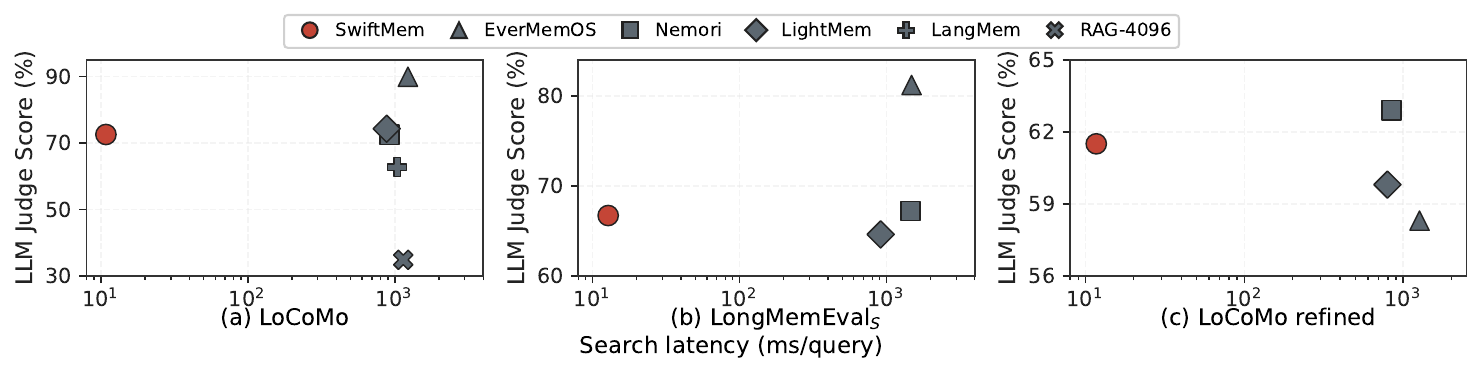}
    % \vspace{-15pt}
    \caption{Overall quality versus query-time search latency (log-scaled $x$-axis). All three panels use the same axis semantics: $x$-axis is search latency and $y$-axis is LLM-judge score. Panels (b) and (c) keep only the four strongest memory baselines for compactness.}
    \label{fig:latency_quality_pair}
\end{figure*}

Figure~\ref{fig:latency_quality_pair} summarizes the overall tradeoff between quality and query-time retrieval latency across the three benchmarks used in this paper.
On LoCoMo, SwiftMem sits at a sharply better latency point while remaining competitive with strong memory systems on LLM-judge score.
On LongMemEval$_S$, reported as a secondary robustness view, the same pattern remains visible: SwiftMem keeps a very large query-time advantage even though EverMemOS is strongest on overall LLM-judge score; the full table and the corresponding scatter plot are reported in Appendix~\ref{app:longmem_results}.
On LoCoMo refined, where the benchmark is recalibrated with stricter LLM judging and cleaned data, Evermemos drops to 58.3\%, SwiftMem remains 61.5\% while preserving a roughly two-order search latency advantage.

\subsection{LoCoMo}
\label{sec:exp_locomo}

\subsubsection{Overall Results}

Tables~\ref{tab:locomo_quality} and~\ref{tab:locomo_latency} report overall LoCoMo quality and cost with GPT-4.1-mini.
EverMemOS has the strongest overall LJ/F1, while SwiftMem remains competitive on LJ and is strongest on B1.
These results show that SwiftMem is not simply trading away quality for latency.

\begin{table*}[t]
\centering
\begin{minipage}[t]{0.48\textwidth}
\centering
\captionof{table}{Overall quality on LoCoMo with GPT-4.1-mini.}
\label{tab:locomo_quality}
\small
\begin{tabular}{lccc}
\toprule
Method & LJ & F1 & B1 \\
\midrule
FullContext & \underline{0.8487} & 0.3848 & 0.3031 \\
RAG-4096 & 0.3481 & 0.2305 & 0.1554 \\
LangMem & 0.6272 & 0.2129 & 0.1368 \\
Nemori & 0.7246 & 0.2036 & 0.1358 \\
LightMem & 0.7427 & 0.2049 & 0.1395 \\
EverMemOS & \textbf{0.8994} & \textbf{0.4731} & \underline{0.3484} \\
SwiftMem (Ours) & 0.7253 & \underline{0.4511} & \textbf{0.4858} \\
\bottomrule
\end{tabular}
\end{minipage}
\hfill
\begin{minipage}[t]{0.48\textwidth}
\centering

\captionof{table}{Add-stage and query-time search cost on LoCoMo with GPT-4.1-mini. Dense retrieval baselines use HNSW-backed vector retrieval.}
\label{tab:locomo_latency}
\footnotesize
\setlength{\tabcolsep}{5pt}
\begin{tabular}{lrr}
\toprule
Method & Add (s) & Search ms \\
\midrule
RAG-4096 & 5460.68 & 1143.604 \\
LangMem & 5561.34 & 1038.770 \\
Nemori & \textbf{4166.25} & 920.507 \\
LightMem & 21448.22 & \underline{881.924} \\
EverMemOS & 11453.53 & 1231.332 \\
SwiftMem (Ours) & \underline{4216.17} & \textbf{10.834} \\
\bottomrule
\end{tabular}
\end{minipage}
\end{table*}

SwiftMem reaches 10.834 ms search latency, while the other strong memory baselines require 881.924--1231.332 ms.
This corresponds to roughly 81--114$\times$ lower query-time search latency than all baselines.
The speedup comes from routing to a smaller query-relevant memory subset, not from comparing against a linear scan.
SwiftMem achieves a significantly lower add time of 4216.17 seconds, demonstrating superior efficiency compared to the tens of thousands of seconds required by LightMem and EverMemos.
The add-stage and query-stage breakdown is in Appendix~\ref{app:breakdown}.

\subsubsection{Category-level Results}
\label{sec:exp_category_qwen}

Table~\ref{tab:locomo_gpt_full} reports category-level LoCoMo results under GPT-4.1-mini.
Overall, SwiftMem improves overlap-oriented metrics while remaining competitive on LLM-judge score against strong memory systems; EverMemOS is strongest on LJ in several categories.
We also observe consistently strong BLEU-1, especially on temporal and single-hop settings as well as overall B1: because SwiftMem routes retrieval to a smaller, query-relevant evidence subset, the returned memories tend to be more lexically concentrated, which better preserves surface n-gram overlap with reference phrasing than diffuse, long-context retrieval.
The results of Open-weight Qwen Model are in Appendix~\ref{app:qwen_results}.

\textbf{Finally, LoCoMo origin is known to mix noisy judging signals with imperfectly dataset cleanness; for a more fair view under stricter judging, we report LoCoMo Refined in \S\ref{sec:exp_locomo_refined}.}

\begin{table*}[t]
\centering
\caption{Category-level LoCoMo results with GPT-4.1-mini.}
\label{tab:locomo_gpt_full}
\scriptsize
\setlength{\tabcolsep}{3.0pt}
\resizebox{\textwidth}{!}{
\begin{tabular}{lccccccccccccccc}
\toprule
& \multicolumn{3}{c}{MH} & \multicolumn{3}{c}{TM} & \multicolumn{3}{c}{OD} & \multicolumn{3}{c}{SH} & \multicolumn{3}{c}{Overall} \\
\cmidrule(lr){2-4}\cmidrule(lr){5-7}\cmidrule(lr){8-10}\cmidrule(lr){11-13}\cmidrule(lr){14-16}
Method & LJ & F1 & B1 & LJ & F1 & B1 & LJ & F1 & B1 & LJ & F1 & B1 & LJ & F1 & B1 \\
\midrule
FullContext & \underline{0.826} & 0.298 & 0.226 & 0.729 & 0.358 & 0.297 & \underline{0.573} & 0.195 & 0.161 & \underline{0.933} & 0.446 & 0.347 & \underline{0.849} & 0.385 & 0.303 \\
Nemori & 0.692 & 0.183 & 0.129 & 0.660 & 0.194 & 0.126 & 0.505 & 0.101 & 0.075 & 0.785 & 0.226 & 0.149 & 0.725 & 0.204 & 0.136 \\
LightMem & 0.663 & 0.199 & 0.144 & \underline{0.732} & 0.106 & 0.073 & 0.510 & 0.109 & 0.076 & 0.800 & 0.256 & 0.171 & 0.743 & 0.205 & 0.140 \\
EverMemOS & \textbf{0.939} & \textbf{0.351} & \underline{0.339} & \textbf{0.809} & \underline{0.472} & \underline{0.465} & \textbf{0.614} & \textbf{0.317} & \underline{0.263} & \textbf{0.952} & \underline{0.473} & \underline{0.348} & \textbf{0.899} & \textbf{0.473} & \underline{0.348} \\
SwiftMem & 0.624 & \underline{0.342} & \textbf{0.393} & 0.704 & \textbf{0.513} & \textbf{0.574} & 0.531 & \underline{0.258} & \textbf{0.273} & 0.789 & \textbf{0.486} & \textbf{0.508} & 0.725 & \underline{0.451} & \textbf{0.486} \\
\bottomrule
\end{tabular}}
\end{table*}

\subsection{LoCoMo Refined}
\label{sec:exp_locomo_refined}

LoCoMo Refined~\cite{locomorefined2026} focuses on two things: making the LLM Judger behave more like a real evaluator, and cleaning the dataset itself so the benchmark becomes more trustworthy.
Table~\ref{tab:locomo_refined} shows that under this stricter protocol, SwiftMem remains strongly competitive on quality: it reaches 61.5, close to Nemori's 62.9 and ahead of both LightMem and EverMemOS.
At the same time, the latency trend is unchanged: SwiftMem still runs at 11.7\,ms/query versus 794.6--1264.4\,ms/query for the other strong memory systems.
\textbf{Notably, the large quality advantage of EverMemOS on original LoCoMo does not persist on LoCoMo refined, whereas SwiftMem's query-time efficiency advantage remains intact.}

For completeness, LongMemEval$_S$ still supports the same qualitative conclusion: SwiftMem retains a large latency advantage with competitive LJ, full results are reported in Appendix~\ref{app:longmem_results}.

\begin{table}[t]
\centering
\caption{Results on LoCoMo Refined.}
\label{tab:locomo_refined}
\small
\begin{tabular}{lrr}
\toprule
Method & LLM Judge Score & Search (ms/query) \\
\midrule
EverMemOS & 0.583 & 1264.4 \\
Nemori & \textbf{0.629} & 843.6 \\
LightMem & 0.598 & \underline{794.6} \\
SwiftMem (Ours) & \underline{0.615} & \textbf{11.7} \\
\bottomrule
\end{tabular}
\end{table}

\begin{figure*}[h!]
    \centering
    \begin{minipage}[t]{0.32\textwidth}
        \centering
        \includegraphics[width=\linewidth]{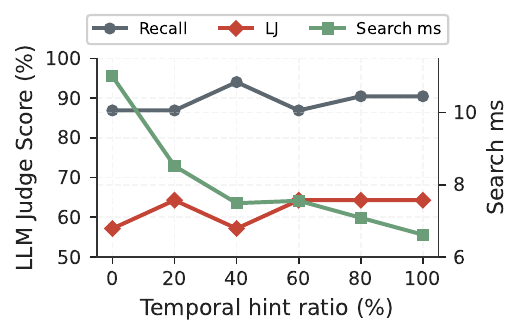}
        \centerline{\small (a) Temporal indexing}
    \end{minipage}
    \hfill
    \begin{minipage}[t]{0.32\textwidth}
        \centering
        \includegraphics[width=\linewidth]{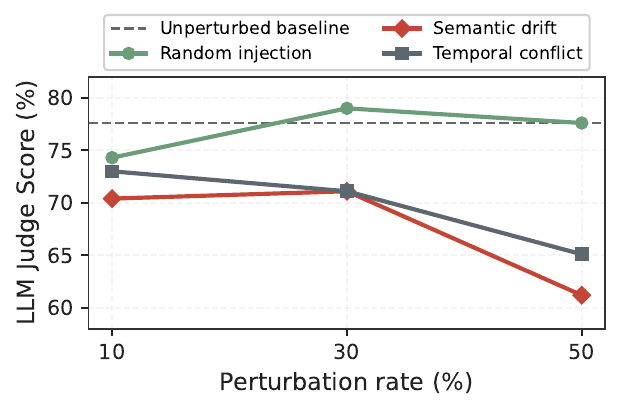}
        \centerline{\small (b) Tag-quality robustness}
    \end{minipage}
    \hfill
    \begin{minipage}[t]{0.32\textwidth}
        \centering
        \includegraphics[width=\linewidth]{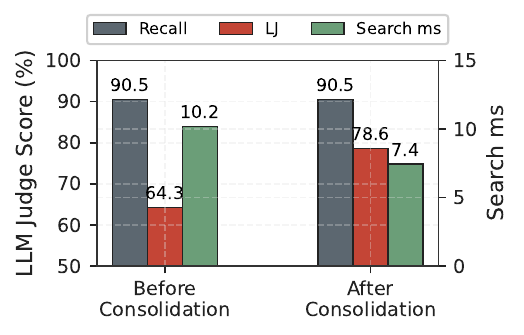}
        \centerline{\small (c) Co-consolidation}
    \end{minipage}
    \caption{Component analyses for SwiftMem. Temporal hints improve latency without harming retrieval quality; tag noise experiments show practical robustness under non-targeted perturbations; and co-consolidation improves answer quality while reducing search latency.}
    \label{fig:component_analyses}
\end{figure*}

\subsection{Ablation Study}
\label{sec:exp_ablation}

We isolate the role of SwiftMem's three main components: temporal indexing, tag quality, and tag-embedding co-consolidation.
Figure~\ref{fig:component_analyses} (a) shows that LLM-judge score remains stable or slightly improved as temporal hints are added, while the corresponding search latency still drops from 11.0\,ms to 6.6\,ms.
Panel (b) shows that SwiftMem is robust to non-targeted perturbations in tag space, while targeted semantic drift and temporal conflict degrade quality more clearly with higher attack rates.
Panel (c) shows that co-consolidation improves LLM-judge score from 64.3 to 78.6; the same setting also reduces search latency from 10.2\,ms to 7.4\,ms.
Detailed experimental setups and complete results are in Appendix~\ref{app:ablation_robustness}.

\section{Conclusion, Limitations and Impact Statement}
\label{sec:conclu}

% We presented SwiftMem, a query-aware agentic memory system that reduces query-time retrieval latency by narrowing the memory search space before ANN-backed vector retrieval.
% By combining temporal indexing, semantic DAG-Tag routing, and embedding-tag co-consolidation, SwiftMem achieves substantially lower search latency than strong memory baselines while preserving competitive answer quality on three benchmarks.
% These results suggest that query-aware indexing is a practical complement to established ANN-based memory systems for scalable, real-time scenarios.

We presented SwiftMem, a query-aware agentic memory system that reduces query-time retrieval latency by narrowing the search space before ANN-backed retrieval.
Across three benchmarks, the results show that query-aware indexing is a practical complement to ANN-based memory systems for scalable, real-time use.

We highlight several scope conditions of the current study.
The temporal-index analysis uses controlled temporal-hint augmentation to stress-test time-aware routing under sparse temporal cues.
SwiftMem also relies on LLM-generated semantic tags, but robustness results show stable behavior under realistic non-adversarial noise and predictable degradation under controlled perturbations.
Finally, SwiftMem is orthogonal to ANN index design: it narrows the query-specific candidate universe, while backends such as HNSW, IVF, or DiskANN determine search efficiency within that reduced space.
Future ANN advances should therefore compose naturally with SwiftMem.

This paper improves long-term memory retrieval efficiency for LLM agents.
Potential benefits include lower latency and compute for memory-augmented applications.
A key risk is easier deployment of persistent memory around sensitive user data.
% \section*{Impact Statement}

% This paper aims to improve the efficiency of long-term memory retrieval for LLM agents.
% Potential positive impacts include lower serving latency and reduced compute for memory-augmented applications.
% Potential risks include making persistent agent memory easier to deploy in settings involving sensitive user data; practical deployments should therefore include consent, retention, and access-control safeguards.

{
\small
\bibliographystyle{plainnat}
\bibliography{example_paper}
}

\newpage
\appendix

\section{Technical Details}

\subsection{Memory Structures}
\label{app:mem_st}

\textbf{Episodic Memory.} It stores individual conversational exchanges as discrete units called \textit{episodes}. Formally, an episode $e_i$ is defined as $e_i = (u_i, m_i, t_i, \mathbf{x}_i)$, where $u_i$ is the user identifier, $m_i$ is the raw conversational content (user utterance and agent response), $t_i$ is the timestamp, and $\mathbf{x}_i \in \mathbb{R}^{d}$ is the semantic embedding. The complete episodic memory is $\mathcal{E} = \{e_1, e_2, ..., e_{N_{mem}}\}$, where $N_{mem}$ denotes the total number of episodes. Episodic memory captures the \textit{what}, \textit{when}, and \textit{who} of conversational interactions.

\textbf{Semantic Memory.} It maintains abstracted, structured knowledge extracted from episodic content. In SwiftMem, semantic memory is realized through the DAG-Tag index $\mathcal{G} = (V, E)$ organizing hierarchical concept relationships, where $V$ represents tag nodes and $E$ represents semantic dependencies. Unlike episodic memory's event-specific nature, semantic memory stores generalized facts and patterns, answering \textit{what is known} about concepts independent of specific conversational instances.

\textbf{Other Memory Types.} 
For example, in Mirix~\cite{wang2025mirixmultiagentmemoryllmbased}, Procedural memory encodes learned behavioral patterns and interaction preferences, such as user communication styles and task-specific workflows. Resource memory maintains references to external artifacts (files, documents) mentioned in conversations. Knowledge vault stores sensitive information (credentials, personal identifiers) with restricted access controls. While SwiftMem focuses primarily on episodic and semantic retrieval, these additional memory types can be incorporated through metadata annotations.

\subsection{Storage and Retrieval}
\label{app:sto_retri}

\textbf{Storage Layer.} Modern agentic memory systems employ a hybrid storage architecture. (1) Vector databases (FAISS~\cite{douze2024faiss}, Milvus~\cite{2021milvus}) store episode embeddings $\{\mathbf{x}_1, ..., \mathbf{x}_{N_{mem}}\}$ enabling similarity-based retrieval through approximate nearest neighbor search, including HNSW, IVF/PQ, or DiskANN-style indexes~\cite{malkov2020hnsw,subramanya2019diskann}. 
(2) Relational databases (PostgreSQL~\cite{Postgres2025}, SQLite~\cite{Sqlite2025}) maintain structured metadata (user IDs, timestamps, tags) supporting filtered queries. 
(3) Graph databases (Neo4j~\cite{neo2025}) store semantic relationships for knowledge graph-based systems. 
% SwiftMem's DAG-Tag structure can be persisted in graph databases or as adjacency structures in relational stores. 
Let $\mathcal{D}$ denote the storage layer; the storage operation for a new episode $e_i$ is $\mathcal{D}.store(e_i) \rightarrow \text{persist}(m_i, \mathbf{x}_i, t_i, u_i, \text{metadata})$.

\textbf{Retrieval Paradigms.} Given a user query $q$ with embedding $\mathbf{e}_q \in \mathbb{R}^{d}$, existing memory frameworks employ several strategies. \textit{Global ANN Search} retrieves nearest episodes from an indexed vector store:
$\mathcal{R}_{global}(q) = \text{top-}k\{\frac{\mathbf{e}_q \cdot \mathbf{x}_i}{||\mathbf{e}_q|| \cdot ||\mathbf{x}_i||}\}_{i=1}^{N_{mem}}$.
This can be sublinear in implementation because the vector store uses ANN indexes, but the searched candidate universe is still the full memory embedding space.
\textit{Filtered Retrieval} applies metadata filters (user ID, time ranges) before or after similarity search:
$\mathcal{R}_{filtered}(q, f) = \text{top-}k\{\frac{\mathbf{e}_q \cdot \mathbf{x}_i}{||\mathbf{e}_q|| \cdot ||\mathbf{x}_i||} \mid f(e_i) = \text{true}\}$.
If filters are not themselves indexed or query-aware, they can still add broad metadata scans or large post-filtering overhead.
\textit{Hybrid Retrieval} combines embeddings, keywords, recency, and graph signals through weighted aggregation, but often keeps the same query-agnostic search scope.
SwiftMem targets this scope-selection gap: it first chooses a temporal or semantic memory subset and then invokes embedding retrieval or reranking only when needed.

% \textbf{The Efficiency Gap.} While some works have explored structured memory retrieval, the critical limitation of existing approaches is their query-agnostic nature; each query always triggers a search across the entire memory storage. 
% As $N_{mem}$ grows linearly with conversation length, retrieval latency scales accordingly, creating unsustainable overhead for long-term agents. 
% SwiftMem addresses this through \textit{query-aware indexing}: analyzing query characteristics (temporal constraints, semantic topics) to identify the relevant memory subset \textit{before} similarity computation, achieving sub-linear complexity.

\subsection{Tag Expansion Example}
\label{sec:tag_ex}

\begin{figure}[h]
    \centering
    \includegraphics[width=0.5\linewidth]{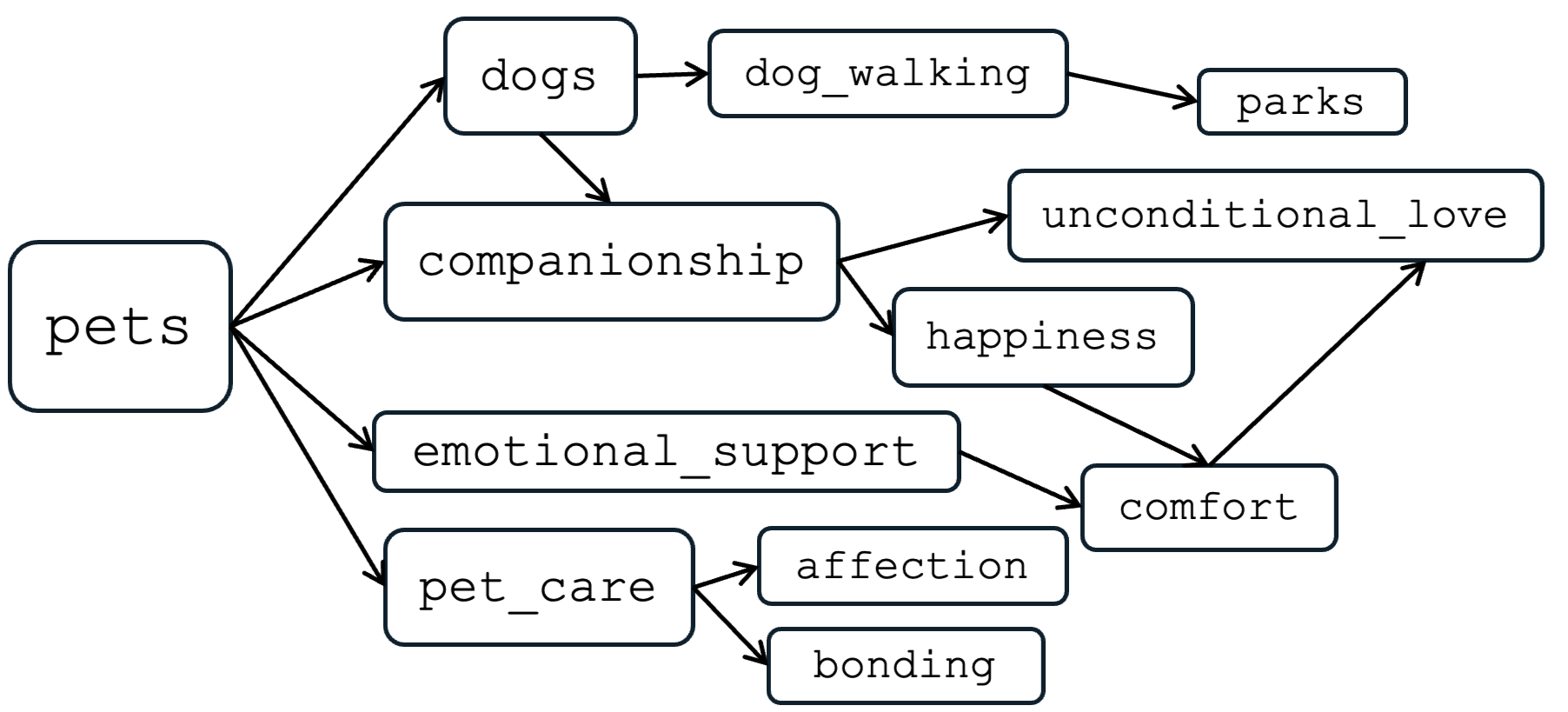}
    \caption{A Semantic DAG-based Tag Example between Andrew and Audrey on LoCoMo~\cite{maharana2024evaluating}. It organizes memories where pets serves as the root topic, with child concepts representing semantic aspects (dogs, companionship) and their derived emotions (happiness, comfort), enabling queries to match either specific terms or abstract themes.}
    \label{fig:dag_toy}
\end{figure}

For example in Figure~\ref{fig:dag_toy}, searching `pets' with depth 2 expands to direct children like `emotional\_support' and `dogs', then to grandchildren such as `dog\_walking' and `unconditional\_love', capturing semantically relevant concepts that facilitate accurate retrieval. This depth-controlled expansion improves search recall by including specific yet relevant concepts while maintaining semantic coherence, making SwiftMem adaptable to different information retrieval scenarios through flexible granularity control.

\subsection{Semantic Specificity Hierarchy Theoretical Analysis}
\label{app:sem_spec}

Theorem~\ref{th:2} formalizes the fundamental property of SwiftMem's hierarchical tag organization: semantic specificity strictly increases from parent to child nodes along any directed path in the DAG. This monotonicity guarantees that broader concepts (e.g., `programming') always precede more specific concepts (e.g., `python\_programming') in the hierarchy, establishing a well-defined semantic ordering. 
SwiftMem dynamically maintains this property through careful DAG construction—when processing new episodes, it creates nodes for new tags, updates episode associations, and establishes parent-child relationships only when the specificity constraint is satisfied.

\subsection{DAG-Tag Query Theoretical Analysis}
\label{app:query}

Theorem~\ref{th:1} establishes SwiftMem's query cost under bounded-frontier DAG-Tag routing. SwiftMem achieves $O(k \cdot (\log |V| + D_{max}))$ query time through two mechanisms: (1) seed-tag lookup in $O(k \cdot \log |V|)$ via a tag-embedding index, and (2) controlled expansion in $O(k \cdot D_{max})$ using DAG adjacency lists, a score-keyed frontier heap, and fixed frontier/candidate budgets. Thus the theorem does not assume unconstrained full-width DAG traversal. Since $k \ll |V| \ll N_{mem}$ and $D_{max} \ll |V|$ in practice (e.g., $k \approx 3$-$10$, $D_{max} \approx 1$-$3$ for typical conversational taxonomies), this represents sub-linear routing relative to memory size before ANN-backed episode retrieval. The space overhead $O(|V| \cdot (d + |\mathcal{P}_{avg}| + |\mathcal{C}_{avg}|))$ for maintaining tag embeddings and DAG adjacency lists grows with semantic vocabulary rather than conversation length, remaining modest even for large-scale deployments. 
The worst case occurs when queries lack semantic specificity and the routing budget is relaxed enough to cover most tags in $V$. In this scenario, the routing advantage diminishes and the system falls back toward broad ANN-backed retrieval, but it does not rely on a linear-scan baseline for its comparison.

\subsection{Implicit Temporal Query and Beyond}
\label{app:temp_im}

Many queries involve temporal reasoning but differ from the explicit `time-to-event' mapping discussed above. 
Instead, they require `event-to-time' mapping, such as an implicit temporal query from LoCoMo~\cite{maharana2024evaluating} beginning with `When':  ``When is Melanie planning on going camping?'' (answer: June 2023).
These queries cannot directly yield time intervals. 
For such cases, we employ the Semantic Tag-DAG method detailed in Section~\ref{sec:dag_tag}.

\textbf{Multi-time-interval Query.}
We also support handling multiple time interval queries for those where several possible time intervals are mentioned.
Existing SOTA works can only process each time range independently, resulting in multiple sequential scans or complex \texttt{UNION} operations. 
In contrast, we first perform intelligent interval merging, which combines overlapping time ranges into a minimal set of non-overlapping intervals. 
% This optimization reduces the number of binary searches required from $O(m \cdot \log n)$ to $O(\log n)$, where $m$ represents the number of original time ranges. 
The merged intervals are then processed on the pre-sorted timeline, ensuring efficient search even with complex temporal query patterns.

\textbf{Temporal Clue Acquisition in Agentic Chat.}
To handle typical queries in real-world applications, we may obtain approximate temporal clues through user-agent interaction. 
Though these temporal clues may be multiple and potentially imprecise, our temporal indexing can also accelerate queries containing such fuzzy temporal information.

\newpage
\subsection{Semantic Tag Clustering Algorithm}
\label{app:clustering}

\paragraph{Clustering Objective.}
Given a tag set $V$ from the DAG-Tag index, our goal is to partition it into cohesive clusters $\{C_1, C_2, \ldots, C_m\}$ that maximize intra-cluster semantic relatedness while respecting hierarchical constraints.

\paragraph{Cohesion Score Definition.}
For a cluster $C = (I_c, V_c, t_c, s_c)$, the cohesion score $s_c$ is computed as:
$s_c = \frac{|E_c|}{|V_c|(|V_c|-1)/2}$
, where $|E_c|$ is the number of edges among tags in $V_c$, and the denominator represents the maximum possible edges in a complete graph of $|V_c|$ nodes.
Higher $s_c$ indicates tighter semantic grouping.

\paragraph{Clustering Algorithm.}
We employ a three-phase approach:

\textbf{1) Hierarchical Grouping:}
Starting from root tags in the DAG, we perform breadth-first traversal to identify subtrees with high edge density.

\textbf{2) Co-occurrence Analysis:}
We build a co-occurrence graph $G_{co} = (V, E_{co})$ where edge $(u, v) \in E_{co}$ exists if tags $u$ and $v$ appear together in $\geq k$ episodes. We merge candidate clusters if their tags have strong co-occurrence links.

\textbf{3) Connected Component Refinement:}
We apply community detection on the combined graph $G_{combined} = (V, E_{DAG} \cup E_{co})$ to identify final clusters.

\subsection{Co-consolidation Implementation Details}
\label{app:consolidation}

\paragraph{Consolidation Triggering.}
We trigger re-consolidation when memory fragmentation exceeds the threshold, estimated by:
$\text{Fragmentation} = 1 - \frac{\sum_i |V_{C_i}| \cdot \text{avg\_emb\_count}_i}{\text{total\_memory\_slots}}$
The process runs asynchronously during low-query periods to minimize impact on retrieval latency.
We prioritize consolidation of clusters with high benefit scores:
$\text{Score}(C_i) = w_1 \cdot s_c + w_2 \cdot \log|V_c| - w_3 \cdot \text{frag}(C_i)$
, where $\text{frag}(C_i)$ measures current fragmentation of cluster $C_i$ embeddings.

\paragraph{Complexity Analysis.}
\textbf{Clustering:} $O(|V|^2)$ for co-occurrence graph construction, $O(|V| \log |V|)$ for community detection;
\textbf{Consolidation:} $O(|V| + |E_{total}|)$ where $|E_{total}|$ is total embeddings across all tags;
\textbf{Amortized Cost:} With periodic re-consolidation (every $N$ queries), per-query overhead is $O(|V|/N) \approx O(1)$.

\paragraph{Memory Layout Optimization.}
Given tag clusters $\{C_1, \ldots, C_m\}$, we reorganize the embedding storage as follows:

\begin{algorithm}[h]
\caption{Co-consolidation Memory Reorganization}
\begin{algorithmic}[1]
\STATE \textbf{Input:} Clusters $\{C_i\}$, current layout $\mathcal{L}$, embedding store $\mathcal{E}$
\STATE \textbf{Output:} Optimized layout $\mathcal{L}'$

\STATE Sort clusters by cohesion score: $C_1, \ldots, C_m$ (descending $s_c$)
\STATE Initialize new memory offset: $o = 0$
\FOR{each cluster $C_i = (I_c, V_c, t_c, s_c)$}
    \FOR{each tag $v \in V_c$}
        \STATE Retrieve embeddings $\{\mathbf{e}_j\}$ associated with $v$
        \STATE Write embeddings to contiguous block starting at $o$
        \STATE Update $\mathcal{L}'(v) = (o, o + |\{\mathbf{e}_j\}|, I_c, |\{\mathbf{e}_j\}|)$
        \STATE $o \leftarrow o + |\{\mathbf{e}_j\}|$
    \ENDFOR
\ENDFOR
\end{algorithmic}
\end{algorithm}

\section{Additional Experiments and Reproducibility Details}
\label{app:extra_exp}

\subsection{Workload Statistics}
\label{app:workload}

Table~\ref{tab:workload} summarizes the retrieval workload measured in our rerun logs.
Temporal queries are a subset of retrieval queries; semantic routing is used for all retrieval queries.

\begin{table}[h]
\centering
\caption{Workload statistics used by the unified rerun.}
\label{tab:workload}
\small
\begin{tabular}{lrrrrr}
\toprule
Dataset & Total & Retrieval & Retrieval Ratio & Temporal & Temporal Ratio \\
\midrule
LoCoMo & 1986 & 1986 & 1.0000 & 91 & 0.0458 \\
LongMemEval$_S$ & 500 & 500 & 1.0000 & 133 & 0.2660 \\
\bottomrule
\end{tabular}
\end{table}

% \subsection{Tag-quality Robustness}
% \label{app:tag_robustness}

% The intrinsic tag-generation quality is stable in the non-adversarial setting: JSON validity is 100\%, fallback rate is 0\%, duplicate raw tags are 0\%, malformed or overly generic raw tags are 4/1155 (0.35\%), and temporal conflicts are 0\%.
% We also evaluate controlled perturbations on a LoCoMo subset.
% Table~\ref{tab:tag_robust_app} shows that random injection has limited impact, while targeted semantic drift and temporal conflict degrade quality more clearly at higher rates.

% \begin{table}[h]
% \centering
% \caption{Tag robustness on a LoCoMo subset. Each cell is LJ/F1/B1.}
% \label{tab:tag_robust_app}
% \small
% \begin{tabular}{lccc}
% \toprule
% Setting & 10\% & 30\% & 50\% \\
% \midrule
% Random injection & 0.743/0.428/0.477 & 0.790/0.443/0.490 & 0.776/0.440/0.486 \\
% Semantic drift & 0.704/0.395/0.438 & 0.711/0.414/0.452 & 0.612/0.348/0.384 \\
% Temporal conflict & 0.730/0.396/0.441 & 0.711/0.407/0.444 & 0.651/0.376/0.427 \\
% \bottomrule
% \end{tabular}
% \end{table}

% The unperturbed baseline for this subset is 0.776/0.451/0.493 (LJ/F1/B1).
% Even at 50\% targeted perturbation, SwiftMem remains functional, suggesting practical robustness rather than brittle tag dependence.

\subsection{Ablation and Robustness Details}
\label{app:ablation_robustness}

\paragraph{Tag-quality Robustness.}
The intrinsic tag-generation quality is stable in the non-adversarial setting: JSON validity is 100\%, fallback rate is 0\%, duplicate raw tags are 0\%, malformed or overly generic raw tags are 4/1155 (0.35\%), and temporal conflicts are 0\%.
We also evaluate controlled perturbations on a LoCoMo subset.
Table~\ref{tab:tag_robust_app} shows that random injection has limited impact, while targeted semantic drift and temporal conflict degrade quality more clearly at higher rates.

\begin{table}[h]
\centering
\caption{Tag robustness on a LoCoMo subset. Each cell is LJ/F1/B1.}
\label{tab:tag_robust_app}
\small
\begin{tabular}{lccc}
\toprule
Setting & 10\% & 30\% & 50\% \\
\midrule
Random injection & 0.743/0.428/0.477 & 0.790/0.443/0.490 & 0.776/0.440/0.486 \\
Semantic drift & 0.704/0.395/0.438 & 0.711/0.414/0.452 & 0.612/0.348/0.384 \\
Temporal conflict & 0.730/0.396/0.441 & 0.711/0.407/0.444 & 0.651/0.376/0.427 \\
\bottomrule
\end{tabular}
\end{table}

The unperturbed baseline for this subset is 0.776/0.451/0.493 (LJ/F1/B1).
Even at 50\% targeted perturbation, SwiftMem remains functional, suggesting practical robustness rather than brittle tag dependence.

\paragraph{Temporal Indexing Ablation.}
We evaluate the improvements brought by temporal indexing on the LongMemEval$_S$ dataset.
In previous evaluations on LoCoMo, cases were too rare for temporal indexing invocation, reflecting a dataset design limitation rather than real-world conversational query patterns, where both explicit and implicit temporal intent are fundamental~\cite{piryani2025itshightimesurvey}.

To better simulate real-world long-term scenarios, we:
1) select a subset of conversations from LongMemEval$_S$ where each contains more than $55$ haystack sessions;
2) add two time intervals (one containing the correct answer and one as the distractor) at varying proportions: 20\%, 40\%, 60\%, 80\%, and 100\% of each question to control the proportion of temporal clues;
3) compare retrieval latency, evidence recall, and end-to-end accuracy between queries with and without temporal hints.

As shown in Figure~\ref{fig:component_analyses}(a), temporal indexing demonstrates significant benefits when temporal hints are available in queries.
Evidence recall keeps stable as the temporal hint ratio increases, while search latency decreases substantially from 11.1 ms to 7.2 ms, achieving a 35\% reduction.
This improvement is attributed to the hierarchical temporal index structure, which allows the system to quickly filter out irrelevant time periods and focus on relevant temporal ranges, thereby reducing computational overhead while maintaining high retrieval quality.

\paragraph{Tag-Embedding Co-consolidation Ablation.}
Figure~\ref{fig:component_analyses}(c) illustrates the impact of tag-embedding co-consolidation on system performance.
Evidence recall remains stable at 90.5\% both before and after consolidation, indicating that the co-consolidation process preserves retrieval quality while optimizing memory structure.
The performance improvement from 64.3\% to 78.6\% under identical evidence recall (90.5\%) demonstrates that consolidation enhances response quality through better memory organization: merging fragmented observations into coherent units reduces context noise and improves the LLM's ability to synthesize evidence, transforming scattered facts into structured narratives that enable more accurate reasoning.
Additionally, search latency decreases from 10.2ms to 7.4ms, validating that tag-embedding co-consolidation effectively addresses memory bloat while improving both response accuracy and computational efficiency.
The accuracy gain occurs because consolidation merges fragmented memories into coherent units, reducing retrieval noise and enabling more consistent LLM responses.

\subsection{Open-weight Model Results}
\label{app:qwen_results}

Table~\ref{tab:qwen_full_app} reports the category-level LoCoMo rerun using Qwen3-30B-A3B-Instruct-2507.
Best per cell item is bold and second best is underlined.

\begin{table}[h]
\centering
\caption{LoCoMo results with Qwen3-30B-A3B-Instruct-2507. Each cell is LJ/F1/B1, with best item bolded and second best underlined.}
\label{tab:qwen_full_app}
\scriptsize
\resizebox{\linewidth}{!}{
\begin{tabular}{lccccc}
\toprule
Method & MH & TM & OD & SH & Overall \\
\midrule
FullContext & \textbf{0.699}/\underline{0.247}/\underline{0.190} & 0.287/\underline{0.171}/\underline{0.124} & 0.312/\underline{0.111}/\underline{0.082} & \textbf{0.883}/\underline{0.361}/\underline{0.263} & \underline{0.690}/\underline{0.285}/\underline{0.209} \\
Nemori & 0.376/0.102/0.076 & 0.200/0.089/0.058 & 0.260/0.064/0.048 & 0.316/0.078/0.053 & 0.299/0.084/0.058 \\
LightMem & \underline{0.677}/0.171/0.121 & \textbf{0.511}/0.105/0.070 & \textbf{0.552}/0.090/0.062 & \underline{0.796}/0.223/0.146 & \textbf{0.699}/0.181/0.120 \\
SwiftMem & 0.638/\textbf{0.304}/\textbf{0.342} & \underline{0.373}/\textbf{0.370}/\textbf{0.402} & \underline{0.354}/\textbf{0.191}/\textbf{0.215} & 0.772/\textbf{0.371}/\textbf{0.379} & 0.612/\textbf{0.347}/\textbf{0.367} \\
\bottomrule
\end{tabular}}
\end{table}

\subsection{LongMemEval$_S$ Results}
\label{app:longmem_results}

LongMemEval$_S$ serves as a secondary robustness benchmark for substantially longer histories and a different task mix.
Under this benchmark, SwiftMem still preserves a large latency advantage and competitive LLM-judge accuracy, while EverMemOS obtains the strongest overall LJ/F1.

\begin{table}[h]
\centering
\caption{Overall quality and search latency on LongMemEval$_S$ with GPT-4.1-mini. Best per column is bold; second best is underlined.}
\label{tab:longmem_app}
\small
\begin{tabular}{lrrrr}
\toprule
Method & LJ & F1 & B1 & Search (ms/query) \\
\midrule
FullContext & 0.634 & 0.260 & 0.220 & 5030.292 \\
Nemori & \underline{0.672} & 0.192 & 0.214 & 1457.153 \\
LightMem & 0.646 & 0.112 & 0.162 & \underline{913.948} \\
EverMemOS & \textbf{0.812} & \textbf{0.386} & \underline{0.294} & 1483.5 \\
SwiftMem (Ours) & 0.667 & \underline{0.314} & \textbf{0.387} & \textbf{12.775} \\
\bottomrule
\end{tabular}
\end{table}

\begin{figure}[h]
\centering
\includegraphics[width=0.5\linewidth]{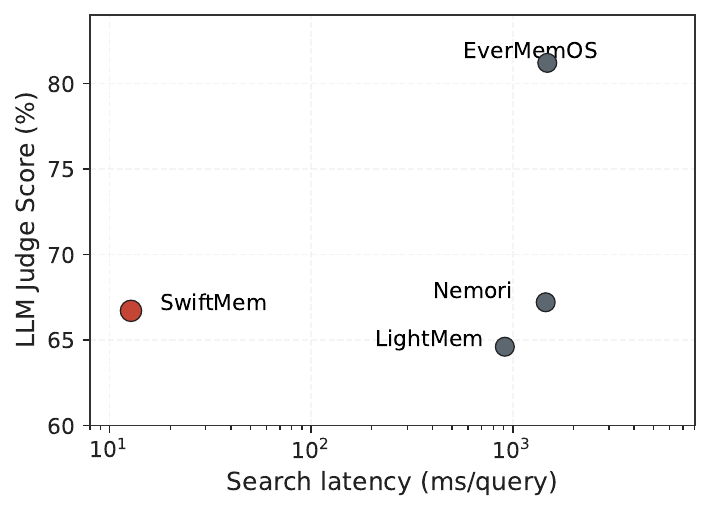}
\caption{LongMemEval$_S$: quality versus query-time search latency. For this secondary benchmark, we keep the strongest four memory systems and report the full numeric comparison in Table~\ref{tab:longmem_app}.}
\label{fig:longmem_app}
\end{figure}

\subsection{Add-stage and Query-stage Breakdown}
\label{app:breakdown}

Table~\ref{tab:add_breakdown_app} decomposes SwiftMem's add-stage overhead.
The expensive components are LLM metadata synthesis, DAG maintenance, and episode insertion/update; co-consolidation is small and can run asynchronously.
More specifically, the add stage is dominated by LLM synthesis and DAG maintenance (2038.093 s) plus episode add/update (2035.076 s), while co-consolidation contributes only 0.705 s on LoCoMo.
This profile supports the intended use case: spend offline or asynchronous work to make repeated online retrieval fast.

\begin{table}[h]
\centering
\caption{SwiftMem add-stage component breakdown on LoCoMo.}
\label{tab:add_breakdown_app}
\small
\begin{tabular}{lrrr}
\toprule
 Model & LLM+DAG (s) & Episode Add/Update (s) & Co-consolidation (s) \\
\midrule
GPT-4.1-mini & 2038.093 & 2035.076 & 0.705 \\
Qwen3-30b-a3b-instruct-2507 & 2118.655 & 1307.668 & 1.166 \\
\bottomrule
\end{tabular}
\end{table}

For query-time search on LoCoMo, SwiftMem averages 1.364 ms for tag inference, 7.395 ms for index search, and 1.285 ms for reranking.
This confirms that the online path is dominated by small bounded-index operations rather than full-memory retrieval.

\subsection{HNSW-backed Baseline Setup}
\label{app:hnsw_setup}

All dense-retrieval memory baselines use the same standard ANN retrieval stack: an HNSW index over \texttt{text-embedding-3-large} vectors with 3072 dimensions and cosine distance. We state this explicitly to clarify the comparison surface: SwiftMem is not being compared against a naive linear scan, but against memory systems already equipped with a strong HNSW-backed vector backend. SwiftMem's gain therefore comes from query-aware candidate narrowing before the ANN-backed retrieval or reranking stage, rather than from replacing the ANN index itself.

\subsection{Reproducibility Resources}
\label{app:reproducibility}

We will provide the supplementary package containing the SwiftMem implementation, baseline wrappers, configuration files, prompts upon acceptance.
The package will include commands for: (1) preparing LoCoMo and LongMemEval$_S$ from their public releases; (2) building HNSW-backed vector indexes for dense-retrieval baselines; (3) running SwiftMem add/search stages; (4) rerunning GPT-4.1-mini and Qwen3-30B-A3B-Instruct-2507 experiments; and (5) reproducing the tables in Section~\ref{sec:exp} and Appendix~\ref{app:extra_exp}.
We will also include the exact prompts used for tag generation.

\subsection{Compute Resources and Assets}
\label{app:compute_assets}

All methods in a comparison are run under the same hardware, model, dataset split, and timing instrumentation.
The experiments require CPU memory sufficient to hold the vector index and metadata stores, GPU or API access for the LLM backbone used in generation/evaluation, and local storage for intermediate memory states and logs.
The supplementary release will list the exact compute workers, memory, storage, software versions, random seeds, wall-clock time per run, and total compute budget used for each reported experiment.

The paper uses public benchmark assets LoCoMo~\cite{maharana2024evaluating} and LongMemEval$_S$~\cite{wulongmemeval}, public or API-accessible LLM backbones including GPT-4.1-mini and Qwen3-30B-A3B-Instruct-2507, and cited baseline systems.
The supplementary package will include asset versions, access instructions, license or terms-of-use notes where available, and documentation for the new SwiftMem code and scripts.
No new crowdsourcing or human-subject data collection is introduced by this work; the datasets are existing public benchmark assets.

\subsection{Tag Generation Prompt}
\label{sec:append}
\begin{tcolorbox}[title = {Tag Generation Prompt}]
You are a semantic tag extraction assistant.
Your task is to:

1. Extract 3-8 meaningful tags that capture the main topics, themes, and contexts

2. Identify hierarchical relationships between these tags (parent-child)

Guidelines for tags:

- Tags should be lowercase, single words or short phrases (max 3 words)

- Focus on: topics, activities, locations, entities, emotions, intents

- Prioritize specific over generic (e.g., `python\_programming' over `technology')

- Use underscores for multi-word tags (e.g., `machine\_learning')

- Avoid overly broad tags like `conversation' or `chat'

Guidelines for relations:

- parent tag = broader/more abstract concept

- child tag = more specific concept

- Only include relations that are clear from the conversation

- Examples:

  * parent: `work', child: `programming'
  
  * parent: `lgbtq', child: `transgender\_story'
  
  * parent: `food', child: `italian\_cuisine'
  
  * parent: `identity', child: `self\_acceptance'

Return ONLY a JSON object:
\begin{mylisting}
{
  "tags": ["tag1", "tag2", "tag3", ...],
  "relations": [
    {"parent": "broader_tag", "child": "specific_tag"},
    ...
  ]
}
\end{mylisting}
If no clear hierarchical relations exist, return an empty `relations' array.
\end{tcolorbox}

% \newpage
% \input{checklist}

\end{document}